\documentclass{article}
\usepackage{spconf,amsmath,graphicx,hyperref}
\usepackage{amsmath}
\usepackage{amsfonts}
\usepackage{multirow}
\usepackage{booktabs}

\title{MOSA: MOTION-GUIDED SEMANTIC ALIGNMENT FOR DYNAMIC SCENE GRAPH GENERATION}
\name{Xuejiao Wang$^{1}$ \qquad Bohao Zhang$^{1}$ \qquad Changbo Wang$^{2}$$^{\star}$\qquad Gaoqi He$^{1}$$^{\star}$ \thanks{${\star}$ denotes the corresponding author.}\thanks{This article is partially supported by Natural Science Foundation of China under Grants 62472178, the Open Projects Program of StateKey Laboratory of Multimodal Artificial Intelligence Systems (No.MAIS2024111), and the Key Technology Research and Development Program Project of the Shanghai Science and Technology Commission under Grants 25511107200.}}
  
\address{$^{1}$ School of Computer Science and Technology, East China Normal University, Shanghai, China \\
$^{2}$School of Data Science and Engineering, East China Normal University, Shanghai, China}
\begin{document}
\ninept
\maketitle
\begin{abstract}
Dynamic Scene Graph Generation (DSGG) aims to structurally model objects and their dynamic interactions in video sequences for high-level semantic understanding. However, existing methods struggle with fine-grained relationship modeling, semantic representation utilization, and the ability to model tail relationships. To address these issues, this paper proposes a motion-guided semantic alignment method for DSGG (MoSA). First, a Motion Feature Extractor (MFE) encodes object-pair motion attributes such as distance, velocity, motion persistence, and directional consistency. Then, these motion attributes are fused with spatial relationship features through the Motion-guided Interaction Module (MIM) to generate motion-aware relationship representations. To further enhance semantic discrimination capabilities, the cross-modal Action Semantic Matching (ASM) mechanism aligns visual relationship features with text embeddings of relationship categories. Finally, a category-weighted loss strategy is introduced to emphasize learning of tail relationships. Extensive and rigorous testing shows that MoSA performs optimally on the Action Genome dataset.
\end{abstract}
\section{Introduction}
\label{sec:intro}
In recent video understanding research~\cite{wang2024human, xi2023chain, xi2023open}, parsing object interactions and fine-grained relationships in dynamic scenes has become central to advancing visual intelligence. Dynamic Scene Graph Generation (DSGG) structures objects and their time-varying relationships in video~\cite{chang2021comprehensive, nguyen2024hig}, supporting higher-order visual reasoning~\cite{wang2022sgeitl,khan2025survey}, video question answering~\cite{cherian20222,mao2022dynamic}, and autonomous driving applications~\cite{liu2024dynamic,zhang2024graphad}. The temporal dynamics, complex object interactions, and subtle action complicate relation modeling~\cite{lin2024td2,wang2023cross}, while long-tailed distributions further hinder the recognition of rare but meaningful relationships.
\begin{figure}[t]
\centering
\includegraphics[width=1.0\linewidth]{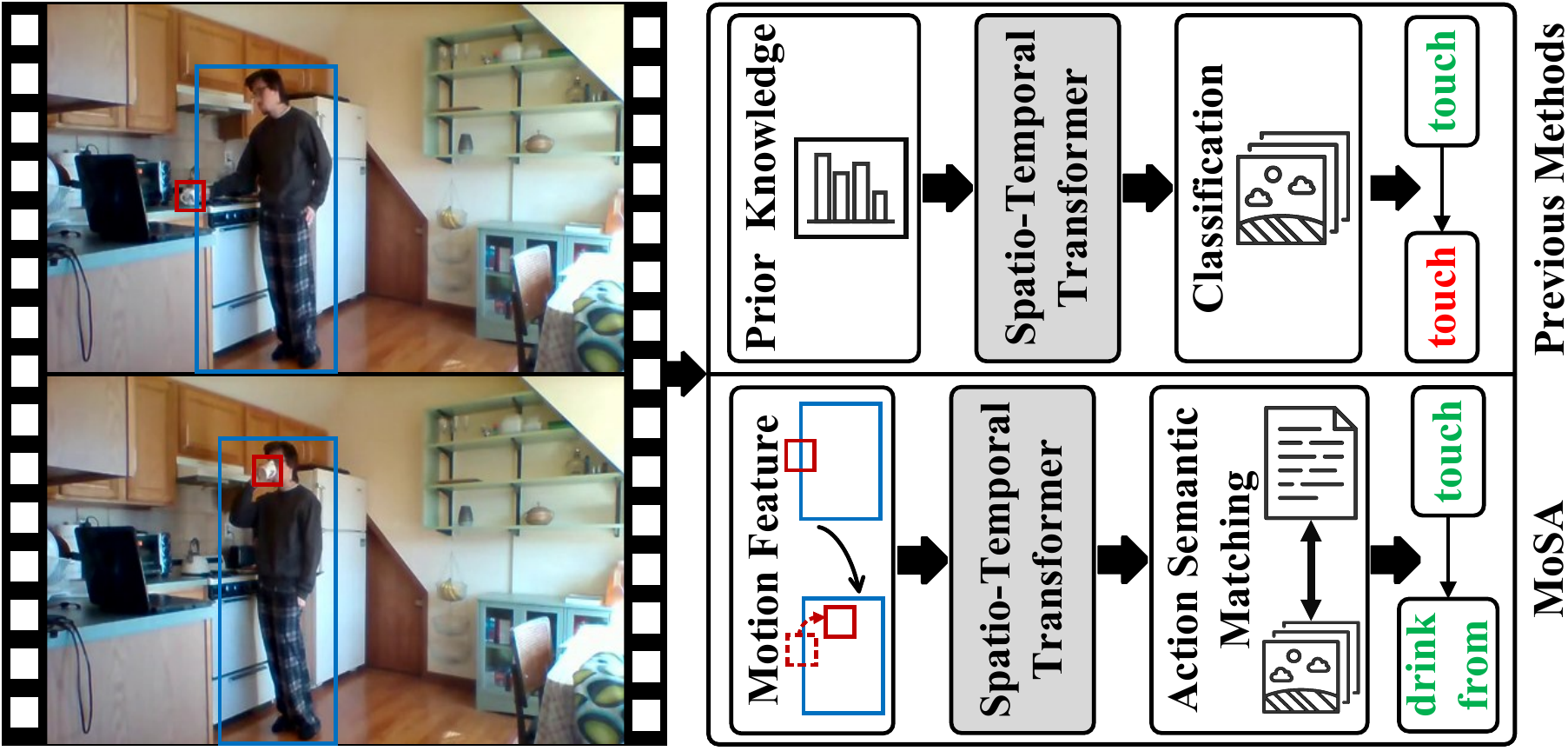}
\vspace{-0.8cm}
\caption{Differences in structure and prediction results between the previous method and MoSA.}
\vspace{-0.5cm}
\label{fig1}
\end{figure}

To handle complex dynamic relationships and the long-tail issue in DSGG, existing research mainly explores two directions: (1) Temporal relationship modeling based on visual features~\cite{cong2021spatial, nag2023unbiased, lin2024td2}, which uses spatio-temporal transformers or graph neural networks to capture cross-frame semantic changes of objects. (2) Methods based on prior knowledge or semantic information enhancement~\cite{li2022dynamic, wang2022dynamic, wang2023cross}, which uses external knowledge bases, co-occurrence statistics, or vision–language pre-trained models to provide semantic guidance or knowledge constraints for relationship prediction.

Although DSGG has progressed, previous methods still struggle with fine-grained actions. They rely on static visual features and coarse-grained temporal context modeling, lacking explicit motion modeling, which hinders distinguishing relationships under different motion patterns. They also ignore semantic prior knowledge of the textual modality, so visual-only information fail to handle semantically similar but motion-different relationships. As shown in Fig.~\ref{fig1}, previous methods depend on prior knowledge and spatio-temporal transformers, confusing the ``drink from'' and ``touch'' relationships. The lack of practical multimodal guidance further weakens fine-grained recognition. Moreover, the long-tail distribution biases models toward high-frequency relationships, leaving low-frequency but meaningful ones ignored and limiting generalization.
\begin{figure*}[t]
\centering
\includegraphics[width=1.0\textwidth]{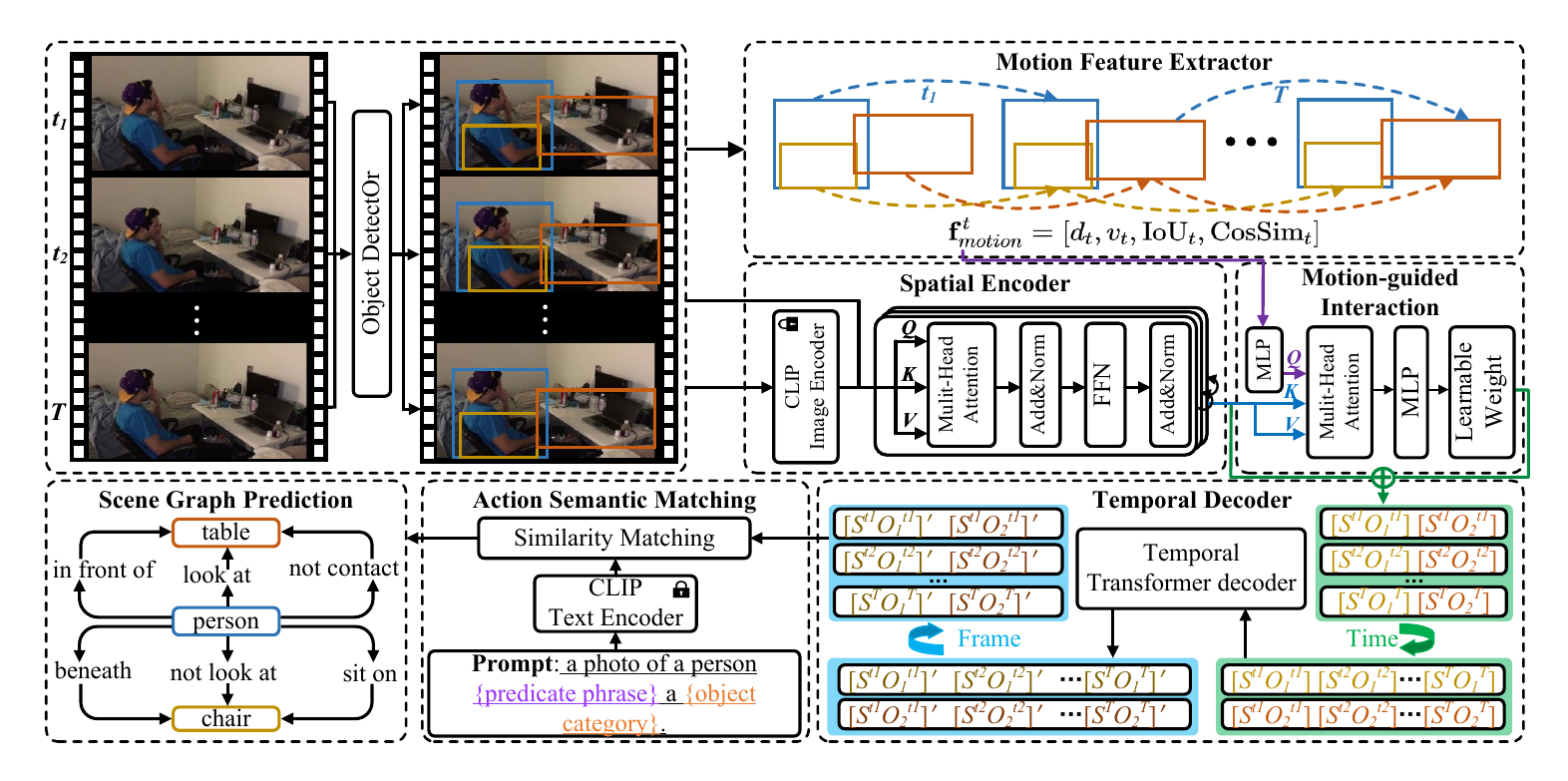} 
\vspace{-1.1cm}
\caption{Overall framework of MoSA. The model detects objects, forms subject–object pairs, and encodes visual–semantic features. Motion features are extracted and fused with spatial features, followed by temporal modeling and semantic alignment for fine-grained DSGG.}
\vspace{-0.4cm}
\label{fig2}
\end{figure*}

To address the above issues, we propose a motion-guided semantic alignment DSGG framework (MoSA). Specifically, a Motion Feature Extractor (MFE) explicitly computes motion attributes between object pairs, which are fused with spatial features in the Motion-guided Interaction Module (MIM) to capture fine-grained action relationships that are difficult to distinguish based solely on appearance. An Action Semantic Matching Module (ASM) aligns visual relationship features with textual embeddings, providing explicit semantic guidance. In addition, a category-weighted loss mitigates long-tail bias and improves generalization.

The main contributions can be summarized as follows. (1) To enhance the ability of DSGG models to recognize complex dynamic relationships, the MoSA architecture is proposed. The MFE module is introduced to model the multidimensional motion attributes between objects explicitly, and the MIM module implements motion-guided relationship feature enhancement to achieve explicit modeling of the interactive relationships between objects. (2) To address the confusion caused by semantically similar relationships, the ASM module is designed. It introduces language priors through cross-modal feature alignment mechanisms, guiding relationship feature learning to achieve stronger semantic distinguishability and enhancing the ability to understand fine-grained semantic relationships. (3) Comprehensive experiments are conducted on the Action Genome dataset. The results show that MoSA not only exhibits superior overall performance but also improves the recognition accuracy of long-tail relationship prediction.

\section{Method}
\subsection{Method Overview}
\textbf{Problem Definition.} DSGG aims to automatically detect objects and their temporal relationships in the input video sequence and construct a structured scene graph to represent multiple objects and their dynamic interaction relationships. Specifically, given a video sequence $V = \{I_1,\dots, I_T\}$, the model detects objects $o_i^t \in \mathcal{O}_t$ in each frame $I_t$ ($t \in [1, T]$) and predicts relationship $r_{ij}^t \in \mathcal{R}$ for each pair of objects $(o_i^t, o_j^t)$. The scene graph is $\mathcal{G}_t = (\mathcal{O}_t, \mathcal{E}_t)$, where $\mathcal{O}_t = \{o_1^t,\dots, o_{N_t}^t\}$ denotes the set of objects in the frame, $\mathcal{E}_t = \{(o_i^t, r_{ij}^t, o_j^t)\}$ denotes the triple of relationships between pairs of objects, $N_t$ is the number of objects in the $t$-th frame, and $\mathcal{R}$ is the set of predefined relationship categories. In the modeling process, DSGG requires models not only to combine static intra-frame relationships but also to combine temporal and motion information to model the dynamic evolution. Its core goal is to maximize the conditional probability distribution of the relationship between each pair of objects.
\begin{equation}
    \hat{r}_{ij}^t=arg \max_{r \in \mathcal{R}}P(r\mid V,o_i^t,o_j^t),
\end{equation}
where $P(r\mid V, o_i^t, o_j^t)$ denotes the posterior probability of relation $r$ given video context $V$ and object pair $(o_i^t, o_j^t)$.

\textbf{Framework.} Fig.~\ref{fig2} shows the overall framework of MoSA. First, the model detects objects in each frame and forms subject–object pairs. For each pair, MoSA fuses object visual features, category semantic embeddings, and global features extracted by the CLIP visual encoder to obtain the initial relationship features. These features are refined by a Spatial Encoder that models global spatial dependencies. The MFE then computes motion attributes (e.g., distance, velocity, IoU, and motion direction consistency) and the MIM fuses them with spatial features to generate motion-informed representations. The Temporal Decoder models their evolution across time, while the ASM aligns relationship features with predicate semantics for relationship prediction.

\subsection{Motion-Driven Relation Modeling}
To model dynamic interaction relationships in videos, we propose a motion-driven relationship encoding mechanism that explicitly models motion attributes to enhance relational features. 

\textbf{Object Detection and Spatial Context Modeling.} For each video frame $I_t$, we use the object detection model~\cite{ren2015faster} to obtain boundary boxes and category labels. Based on detections, we generate candidate pairs $\mathcal{P}_t = \{(o_i^t, o_j^t)\}$ using a $<$person-object$>$ strategy. For each pair, we fuse multi-level spatial features with category embeddings to obtain the initial relationship feature $\mathbf{f}_{rel}$. To complement high-level visual expressions, we use CLIP~\cite{radford2021learning} to encode the union region, obtaining the global visual feature $\mathbf{f}_{clip}$. The concatenated pair feature $\mathbf{f}_{pair}=[\mathbf{f}_{rel},\mathbf{f}_{clip}]$ fuses spatial and visual information for the subsequent Spatial Encoder.

To capture spatial dependencies between objects, all $\mathbf{f}_{pair}$ are organized into a sequence $\mathbf{X}^t$ and input into the spatial Transformer $\mathcal{T}_s$, which performs multi-layer spatial self-attention encoding:
\begin{equation}
   \mathbf{F}_{\text{spatial}}^t=\mathcal{T}_s(\mathbf{X}^t),
\end{equation}
where $\mathbf{F}_{\text{spatial}}$ is the updated pair feature with spatial, semantic, and visual context.

\textbf{Motion-aware Relation Feature Enhancement.} Although the spatial encoder captures spatial and semantic dependencies between objects, spatial information alone cannot fully represent dynamic interaction features in videos. To address this limitation, we propose a motion-aware relationship feature enhancement mechanism. Specifically, the MFE models motion attributes of each object pair $(o_i, o_j)$. Based on the bounding box coordinates, we calculate the center point positions $\mathbf{c}_i^t$ and $\mathbf{c}_j^t$ of the subject-object pair at frame $t$, and then compute the relative distance $d_t$ between objects, which reflects their spatial interaction trend in each frame:
\begin{equation}
    d_{t} =\left \| \mathbf{c}_i^t-\mathbf{c}_j^t \right \| _2.
\end{equation}

Next, we calculate the approach velocity $v_t$ to capture how this distance changes over time, reflecting whether objects are moving closer or farther apart:
\begin{equation}
   v_t=\frac{d_t-d_{t-1}}{\bigtriangleup t},
\end{equation}
where $\bigtriangleup t$ denotes the frame interval. To evaluate the stability of positions, we introduce the mean IoU of sliding window $\mathrm{IoU}_t$ over $K$ frames, reflecting the continuity or randomness of the relationships between objects.
\begin{equation}
   IoU_t=\frac{1}{K} \sum_{k=1}^{K} IoU(b_{i}^{t-k},b_{j}^{t-k}),
\end{equation}
where $b_{i}^{t-k}$ and $b_{j}^{t-k}$ denote the bounding boxes of the subject and object of the $i$th object pair at frame $t-k$, respectively, and $\mathrm{IoU}(\cdot,\cdot)$ is the standard intersection-over-union function. Finally, we compute motion direction consistency $\mathrm{CosSim}_t$, which reflects the coordination of objects movements.
\begin{equation}
   \mathrm{CosSim}_t=\frac{(c_i^t-c_i^{t-1})\cdot (c_j^t-c_j^{t-1})}{\left \| c_i^t-c_i^{t-1} \right \|_2 \left \| c_j^t-c_j^{t-1} \right \|_2}.
\end{equation}

The four attributes are concatenated and passed through MLP to generate high-level motion features $\mathbf{F}_{motion}^t$. The MIM then fuses motion and spatial features with an attention mechanism. Here, motion features provide dynamic guidance, while spatial features $\mathbf{F}_{\text{spatial}}^t$ serve as contextual input. This process generates relationship prompt features with dynamic perception capabilities.
\begin{equation}
    \mathbf{F}^t_{\text{mim}}=MIM(\mathbf{F}_{motion}^t ,\mathbf{F}_{\text{spatial}}^t).
\end{equation}

Finally, we fuse motion-informed features $\mathbf{F}_{\text{mim}}^t$ with the original spatial features $\mathbf{F}_{\text{spatial}}^t$ through residual connections, generating joint relationship features that capture both semantic structure and dynamic motion patterns.
\begin{equation}
\mathbf{F}^{t}_{\text{joint}}=\mathbf{F}^{t}_{\text{spatial}}+\mathbf{F}^{t}_{\text{mim}}.
\end{equation}

\subsection{Temporal-Semantic Relation Reasoning}
\textbf{Temporal Decoder.} To capture temporal dynamics of object relationships, we reorganize the $\mathbf{F}^{t}_{\text{joint}}$ by object pairs and temporal order, forming trajectories $\mathcal{S}(i,j)=[\mathbf{F}
_{\text{joint}}^1(i,j),\ldots,\mathbf{F}_{\text{joint}}^T(i,j)]$ across the video~\cite{wang2023cross}. This sequence is input into a multi-layer temporal Transformer decoder $\mathcal{T}_{\text{t}}$ for global temporal modeling, learning both dynamic evolution and long-term dependencies.
\begin{equation}
    \mathbf{F}_{\text{temporal}}^t=\mathcal{T}_t(\mathcal{S}{(i,j)}),
\end{equation}
where $\mathbf{F}_{\text{temporal}}^t$ is the contextual relationship feature of the pair. This module captures the complex temporal variations of the relationship features, providing a foundation for DSGG.

\textbf{Action Semantic Matching.} To better capture complex interactions, we introduce ASM in the classification stage, which incorporates textual semantic prior knowledge. Textual descriptions are generated for all subject-relationship-object pairs from the dataset category sets. The CLIP text encoder extracts its semantic embeddings to construct a matrix $\mathcal{Z} \in \mathbb{R}^{Nr\times D}$, where $Nr$ is the number of relationship categories and $D$ is the feature dimension. For each object pair $(o_i, o_j)$ at frame $t$, with relationship feature $\mathbf{F}_{\text{temporal}}^t{(i,j)} \in \mathbb{R}^D$, ASM computes dot-product similarity with all text embeddings to obtain cross-modal matching scores.
\begin{equation}
    \mathbf{s}^{t}_{ij}=\mathcal{F}_{\text{temporal}}^t{(i,j)}\cdot \mathbf{Z}^\top \in \mathbb{R}^{Nr}.
\end{equation}

For multi-label relationship categories, Sigmoid activation produces the probability distribution $\hat{p}_{ij}^t= \sigma (\mathbf{s}^{t}_{ij})$.

\subsection{Loss Function}
The overall loss consists of object detection loss and relationship prediction loss. The object detection loss $\mathcal{L}_{\mathrm{obj}}$ uses cross-entropy for object classification. To address category imbalance in relationships, we introduce category-weighted loss, where the weight $\alpha_r$ is calculated by the frequency $n_r$ of category $r$.
\begin{equation}
    \alpha _r=\frac{1/\log n_r}{\frac{1}{Nr} {\textstyle \sum_{k=1}^{Nr}1/\log n_k }}. 
\end{equation}
The relationship loss is then defined as: 
\begin{equation}
    \mathcal{L}_{\text {rel}}=\frac{1}{B} \sum_{i=1}^{B} \alpha_{g_{i}} \cdot\left(1-\hat{p}_{i}\right)^{\gamma} \cdot \mathcal{L}_{\mathrm{BCE}}\left(\mathbf{p}_{i}, \mathbf{y}_{i}\right),
\end{equation}
where $B$ is the batch size, $\mathbf{p}_{i}$ and $\mathbf{y}_{i}$ are predictions and labels, $g_{i}$ is the category label corresponding to the sample, and $\gamma$ is a hyperparameter. The total loss is $\mathcal{L}=\mathcal{L}_{obj}+\mathcal{L}_{rel}$.

\section{Experiments}
\begin{table*}[t]
\centering
\caption{Comparison of Recall@K (R@\{10/20/50\}) performance of PREDCLS, SGCLS, and SGDET tasks on the AG dataset. The best results are marked in \textbf{bold}, and the second-best result is \underline{underlined}.}
\resizebox{\textwidth}{!}
{\begin{tabular}{c|ccc|ccc}
 \toprule
{\multirow{3}{*}{Model}} &  \multicolumn{3}{c|}{With Constraint}             & \multicolumn{3}{c}{No Constraint}                \\ \cline{2-7} 
                       & PREDCLS        & SGCLS          & SGDET          & PREDCLS        & SGCLS          & SGDET          \\
                       & R@\{10/20/50\} & R@\{10/20/50\} & R@\{10/20/50\} & R@\{10/20/50\} & R@\{10/20/50\} & R@\{10/20/50\} \\ \hline
                       RelDN~\cite{Zhang_2019_CVPR} &  66.3/69.5/69.5     &  44.3/45.4/45.4  &  24.5/32.8/34.9  &   75.7/93.0/99.0  &   52.9/62.4/65.1    &  24.1/35.4/46.8  \\
                       GPS-Net~\cite{lin2020gps}&  66.8/69.9/69.9    & 45.3/46.5/46.5   &   24.7/33.1/35.1   & 76.0/93.6/99.5   &  53.6/63.3/66.0  & 24.4/35.7/47.3\\
                       TRACE~\cite{teng2021target}& 64.4/70.5/70.5&36.2/37.4/37.4&19.4/30.5/34.1&73.3/93.0/99.5&36.3/45.5/51.8&27.5/36.7/47.5\\
                       STTran~\cite{cong2021spatial} & 68.6/71.8/71.8 & 46.4/47.5/47.5&25.2/34.1/37.0&77.9/94.2/99.1& 54.0/63.7/66.4& 24.6/36.2/48.8\\
                       STTran-TPI~\cite{wang2022dynamic}&69.7/72.6/72.6&47.2/48.3/48.3&26.2/34.6/37.4&-/-/-&-/-/-&-/-/-\\                       TEMPURA~\cite{nag2023unbiased}&68.8/71.5/71.5&47.2/48.3/48.3&\underline{28.1}/33.4/34.9&80.4/94.2/99.4&56.3/64.7/\underline{67.9}&\underline{29.8}/38.1/46.4\\
                       TD$^2$-Net~\cite{lin2024td2}&\underline{70.1}/-/\underline{73.1}&\textbf{51.1}/-/\textbf{52.1}&\textbf{28.7}/-/37.1&\underline{81.7}/-/\underline{99.8}&\textbf{57.2}/-/\textbf{69.8}&\textbf{30.5}/-/49.3\\
                       MoSA&\textbf{70.6/73.5/73.5}&\underline{48.1}/\underline{49.1}/\underline{49.1} & 26.7/\textbf{35.5}/\textbf{38.3} &\textbf{82.8/96.5/99.9}& \textbf{57.2}/\underline{64.7}/66.8 &27.6/\textbf{39.2}/\textbf{50.1}\\
\toprule
\end{tabular}}\vspace{-0.6cm}
\label{table1}
\end{table*}

\begin{table*}[t]
\centering
\caption{Comparison of mean Recall@K (mR@\{10/20/50\}) performance of PREDCLS, SGCLS, and SGDET tasks on the AG dataset. The best results are marked in \textbf{bold}, and the second-best result is \underline{underlined}.}
\resizebox{\textwidth}{!}
{\begin{tabular}{c|ccc|ccc}
 \toprule
{\multirow{3}{*}{Model}} &  \multicolumn{3}{c|}{With Constraint}             & \multicolumn{3}{c}{No Constraint}                \\ \cline{2-7} 
                       & PREDCLS        & SGCLS          & SGDET          & PREDCLS        & SGCLS          & SGDET          \\
                       & mR@\{10/20/50\} & mR@\{10/20/50\} & mR@\{10/20/50\} & mR@\{10/20/50\} & mR@\{10/20/50\} & mR@\{10/20/50\} \\ \hline
                       RelDN&  6.2/6.2/6.2     &  3.4/3.4/3.4  &  3.3/3.3/3.3  &  31.2/63.1/75.5  & 18.6/36.9/42.6  & 7.5/18.8/33.7 \\
                       TRACE& 15.2/15.2/15.2&8.9/8.9/8.9&8.2/8.2/8.2&50.9/73.6/82.7&31.9/42.7/46.3&\underline{22.8}/\textbf{31.3}/41.8\\
                       STTran& 37.8/40.1/40.2 & 27.3/28.0/28.0&16.6/20.8/22.2&51.4/67.7/82.7& 40.7/50.1/58.8& 20.9/29.7/39.2\\
                       STTran-TPI&37.3/40.6/40.6&28.3/29.3/29.3&15.6/20.2/21.8&-/-/-&-/-/-&-/-/-\\                       
                       TD$^2$-Net&\underline{41.9}/-/\underline{44.8}&\textbf{33.9}/-/\textbf{34.9}&\underline{17.2}/-/22.3&\textbf{61.0}/-/96.4&\textbf{50.1}/-/\textbf{67.9}&\textbf{23.2}/-/\underline{42.1}\\
                       MoSA&\textbf{44.3/47.7/47.8} &\underline{33.0}/\underline{34.2}/\underline{34.2} & \textbf{17.6}/\textbf{23.3}/\textbf{25.2}&\underline{59.9}/\textbf{84.8/98.9} &\underline{45.2}/\textbf{59.3}/\underline{65.0} &19.9/\underline{30.6}/\textbf{45.1}\\
\toprule
\end{tabular}}\vspace{-0.5cm}
\label{table2}
\end{table*}
\subsection{Experimental Setting}
To evaluate the effectiveness of MoSA on DSGG, we conducted experiments on the Action Genome (AG)~\cite{ji2020action} dataset under three tasks: Predicate Classification (PREDCLS), Scene Graph Classification (SGCLS), and Scene Graph Detection (SGDET). Specifically, the PREDCLS task provides the model with the ground-truth bounding boxes and categories of entities and requires the prediction of semantic relationships between all pairs of entities. The SGCLS task provides the ground-truth bounding boxes and requires the model to predict both entity categories and semantic relationships between entities. The SGDET task provides only the raw video frames, and the model needs to automatically detect all entities in each frame, predict their bounding boxes and categories, and infer paired relationships.  We adopt Recall@K (R@\{10/20/50\}) and mean Recall@K (mR@\{10/20/50\}) as metrics, where mR@K better reflects performance on long-tailed categories. All experiments were implemented in PyTorch with Faster R-CNN (ResNet101 backbone)~\cite{ren2015faster, he2016deep} for object detection and CLIP ViT-B/32~\cite{radford2021learning} as the text–image encoder, using Adam with a learning rate $1\times10^{-5}$.

\subsection{Comparison with SOTA Methods}
\begin{figure}[t]
\centering
\includegraphics[width=1.0\columnwidth]{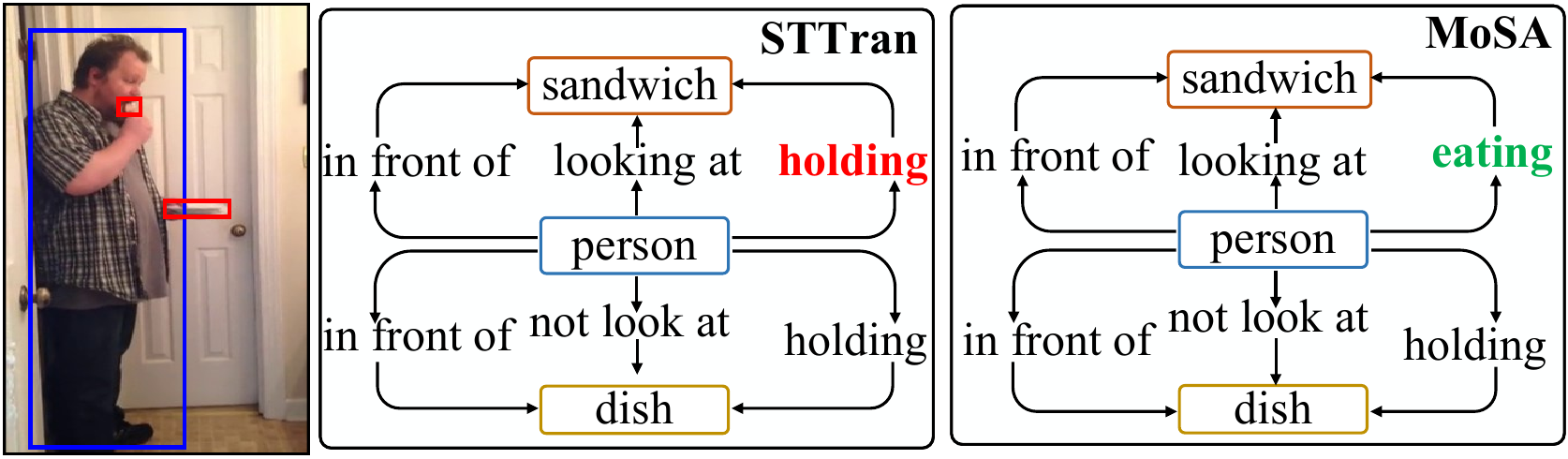}
\vspace{-0.9cm}
\caption{Comparative qualitative results of MoSA and STTran~\cite{cong2021spatial} on the AG dataset.}
\vspace{-0.25cm}
\label{fig3}
\end{figure}

As shown in Table~\ref{table1}, MoSA outperforms mainstream DSGG methods under both With Constraint and No Constraint settings. In PREDCLS, it reaches 70.6\% and 73.5\% at R@10 and R@50, surpassing the current SOTA model TD$^2$-Net~\cite{lin2024td2}. In SGDET, MoSA also performs competitively, achieving 38.3\% at R@50. Table~\ref{table2} further shows improvements in mR@K, with MoSA raising mR@50 in SGDET by 2.9\% and 3.0\% under the two settings, confirming its advantage on long-tailed relations. Overall, MoSA performs well across all tasks, with particularly notable advantages in PREDCLS. Fig.~\ref{fig3} illustrates a qualitative example where STTran predicts the coarse relation ``holding'' for $<$person, sandwich$>$, while MoSA correctly identifies the fine-grained action ``eating'', showing its ability to capture motion-guided semantics better.

\begin{table}[t]
\centering
\caption{Ablation study on key components of MoSA. All results are evaluated under the With Constraint setting. ``w/o'' denotes without.}
\resizebox{\columnwidth}{!}
{\begin{tabular}{c|c|c|c|c}
\toprule
\multirow{2}{*}{Exp}&\multirow{2}{*}{Module} & PREDCLS & SGCLS & SGDET                                           \\ \cline{3-5} 
                        && R@\{10/20/50\} & R@\{10/20/50\} & R@\{10/20/50\}\\ \hline
                   1&\multicolumn{1}{l|}{w/o MFE}&{68.9/71.9/71.9} & {46.8/47.6/ 47.6} & {25.9/34.5/37.3}\\
                   2&\multicolumn{1}{l|}{w/o MIM}& {70.3/73.1/73.1} & {47.9/48.9/ 48.9} &{26.5/35.2/38.0} \\
                   3&\multicolumn{1}{l|}{w/o ASM}&{69.9/72.6/72.7}& {47.7/48.6/ 48.6} &{26.3/35.1/37.8}\\
                   4&\multicolumn{1}{l|}{MoSA} & {70.6/73.5/73.5} &{48.1/49.1/ 49.1}  & {26.7/35.5/38.3} \\
\toprule
\end{tabular}}
\vspace{-0.35cm}
\label{table3}
\end{table}

\subsection{Ablation Study}
To evaluate the contribution of each module, we designed ablation experiments by removing MFE, MIM, and ASM individually on the AG dataset under the With Constraint setting, using R@\{10/20/50\} as metrics (Table~\ref{table3}). Removing MFE produces the largest performance drop, especially in the PREDCLS task, since the model can no longer capture dynamic motion attributes. Because MIM relies on motion features from MFE, this setting also disables MIM, which further exacerbates the performance decline. When MIM alone is removed, motion and spatial features are fused only by concatenation rather than attention, leading to a small decrease. For instance, R@50 in SGCLS decreases from 49.1\% to 48.9\%, showing that attention-based fusion more effectively fusing and mining multimodal information. Removing ASM also reduces performance, with R@10 in PREDCLS dropping by 0.7\% and R@50 in SGDET decreasing by 0.5\%. Overall, each module contributes positively, with MFE and MIM providing the largest gains.

\section{Conclusion}
We propose MoSA, a motion-aware and semantics-aligned method for DSGG. This method explicitly models the multidimensional motion attributes between object pairs. It integrates motion information with spatial relationship features through a motion-guided interaction mechanism, thereby achieving precise modeling of fine-grained dynamic relationships in videos. Additionally, we introduce an action semantic matching module to align visual relationship features with language embeddings of predefined relationship categories, significantly enhancing the ability of the model to distinguish long-tail relationships. Extensive experiments on the AG dataset demonstrate that MoSA performs well in the PREDCLS, SGCLS, and SGDET tasks under both With Constraint and No Constraint settings. Ablation experiments further validate the critical role of each core module in the performance of MoSA, and qualitative results also indicate that MoSA demonstrates superior performance in fine-grained action relationship prediction.

\bibliographystyle{IEEEbib}
\bibliography{refs}

\end{document}